\documentclass{article} 
\usepackage{iclr2027_conference, times}
\iclrfinalcopy

\usepackage{amsmath,amsfonts,bm}

\def\eqref#1{equation~\ref{#1}}

\def\1{\bm{1}}

\DeclareMathAlphabet{\mathsfit}{\encodingdefault}{\sfdefault}{m}{sl}
\SetMathAlphabet{\mathsfit}{bold}{\encodingdefault}{\sfdefault}{bx}{n}

\usepackage{graphicx}
\usepackage{tikz}
\usetikzlibrary{positioning}
\usepackage{forest}
\usepackage{multirow}
\usepackage{booktabs}
\usepackage{caption}
\usepackage{wrapfig}
\usepackage{hyperref}
\hypersetup{
    colorlinks=true,
    citecolor=blue,  
    linkcolor=blue,  
    urlcolor=blue    
}
\usepackage{algorithm}
\usepackage{algorithmic}
\usepackage{url}
\definecolor{myblue}{RGB}{80,150,255}
\definecolor{mygreen}{RGB}{100,200,120}
\definecolor{myyellow}{RGB}{255,210,80}
\definecolor{myred}{RGB}{240,100,100}

\title{
\parbox{\textwidth}{
\centering
Feedback Makes Perfect: A Closed-Loop
Framework for NL-to-STL Translation
}
}

\author{
\makebox[\textwidth][c]{%
\begin{tabular}{c}
\\[-0.2em]
\textbf{Bowen Ye} \qquad \textbf{Xiang Yin} \\[0.7em]
Shanghai Jiao Tong University \\
Shanghai, China \\[0.6em]
\texttt{\{yebowen1025,yinxiang\}@sjtu.edu.cn}
\\[-0.6em]
\end{tabular}
}
}

\begin{document}

\maketitle

\begin{abstract}
Signal Temporal Logic (STL) enables rigorous verification and control of cyber-physical systems, but writing correct specifications requires expertise that most requirement holders lack. Large language models can translate natural-language (NL) requirements into STL, yet stronger translators alone approach an accuracy ceiling. We argue that this ceiling stems from how the task is posed: one-shot, open-loop translation is somewhat ill-defined. Natural language is ambiguous, and, more fundamentally, what a person writes may not always be what they intend, so the target specification is not fully contained in the input text. We therefore reformulate NL-to-STL translation as a closed-loop feedback process. Each generated formula is translated back into natural language for the user to check, and natural-language corrections drive revision until the user accepts the specification. Users never read or write formal syntax. This framework rests on an asymmetry familiar from feedback control theory. The forward path, from ambiguous language to formal logic, is hard and error-prone. The feedback path, from structured STL back to language, can be made highly precise, and a precise feedback path lets an imprecise forward path achieve precise closed-loop behavior. Experiments on 500 expert-authored requirements and seven LLMs support this view. Back-translated explanations agree with expert judgments in 99.5\% of cases. Closed-loop refinement raises strong models from about 89\% open-loop accuracy to 98.0--99.2\%, and yields gains of over 30 percentage points for weaker models (e.g., 17.6\%$\rightarrow$48.0\%). Ablations show these gains come from the semantic content of the feedback rather than from repeated attempts. An expert audit and a 280-session user study further confirm the reliability of the loop. We also identify a capability threshold above which feedback no longer helps.

\end{abstract}
\section{Introduction}
Formal specifications provide a foundation for trustworthy autonomous systems by enabling rigorous verification, runtime monitoring, and controller synthesis. In safety-critical cyber-physical systems, Signal Temporal Logic (STL) offers a precise formalism for expressing temporal behaviors over real-valued signals, including safety constraints, response requirements, and complex temporal objectives~\citep{alur2015principles,maler2004monitoring}. Such specifications have been widely adopted in autonomous driving~\citep{arechiga2019specifying,mehdipour2023formal}, robotics~\citep{silano2021power}, and formal controller synthesis~\citep{belta2019formal,yin2024formal}. Writing them, however, remains an expert task. A correct STL formula requires fluency in temporal operators, interval semantics, and quantitative predicates. The domain engineers and end users who actually hold the requirements rarely have this expertise. Formal guarantees are therefore only as reliable as the translation from what a person wants to what the formula states, and this translation has traditionally depended on a small pool of specialists.  

Recent advances in large language models (LLMs) offer a way around this bottleneck by translating natural-language requirements directly into temporal logic. Substantial progress has been made through dedicated datasets and translation models~\citep{he2022deepstl,chen2023nl2tl,mao2024nl2stl}, external knowledge augmentation~\citep{fang2025enhancing}, reinforcement learning with task-specific rewards~\citep{fang2026restl}, and tool-assisted reasoning~\citep{mendoza2024translating,ye2026reasonstl}. Most of these advances strengthen the translator itself. Yet stronger translators alone appear to approach a ceiling. In our experiments, frontier-scale models plateau at roughly 89\% one-shot accuracy, and a newer model generation yields no further gain. We argue that this ceiling reflects how the task is posed rather than a lack of model capability. One-shot, open-loop translation is ill-defined for two reasons. First, natural language is inherently ambiguous. A requirement can admit several formalizations, and without further information no translator can reliably select the intended one.  Second, and more fundamentally, the words a person writes may not be the intent they hold. Requirements routinely omit thresholds, leave triggering conditions implicit, or compress temporal scope, and their authors would immediately correct these gaps upon seeing the consequences. 
The target of translation is therefore not fully contained in the input text. No amount of translator capability can recover information that was never written down.

\begin{figure}[t]
    \centering
    \includegraphics[width=0.90\linewidth]{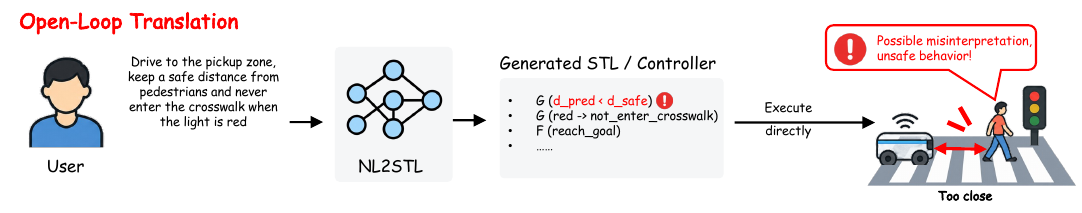}
    \vspace{-1em}
    \caption{One-shot translation.}
    \vspace{-2em}
    \label{fig:motivation}
\end{figure}
Human collaboration offers a familiar resolution. When a supervisor assigns a task, a capable engineer does not act on a first reading. Instead, they restate their understanding, have it confirmed or corrected, and proceed only once the two agree. Requirement clarification methods~\citep{cosler2023nl2spec,fang2025clarifystl} capture part of this exchange by resolving ambiguity in the input before translation. However, they cannot detect a well-posed requirement that has been translated into a formula that says something else. Following this observation, we formulate NL-to-STL translation as a \textbf{closed-loop feedback process} rather than a one-shot task (Fig.~\ref{fig:motivation}). Each generated formula is rendered back into natural language and checked against the author's intent. If it does not match, it is revised using natural-language corrections, yielding the cycle ``$\text{Generate} \rightarrow \text{Explain} \rightarrow \text{Check} \rightarrow \text{Feedback} \rightarrow \text{Revise}$''. Users interact only in natural language, and STL formulas remain entirely in the machine layer.

This formulation brings a classical insight from \emph{feedback control theory} to specification synthesis: a precise feedback path can make an imprecise forward path yield precise closed-loop behavior. In a negative-feedback amplifier, the closed-loop gain is set by the feedback network and becomes largely insensitive to variations in the forward gain. Our framework shares this structure. The forward path, from natural language to STL, is hard and error-prone because it must map complex, often underspecified human intentions into a structured formalism. The feedback path, from STL back to natural language, is far easier. It starts from a structured, unambiguous artifact whose predicates, thresholds, intervals, and logical dependencies can be read off directly and rendered faithfully. This asymmetry is the key to our approach: \emph{we do not need a perfect translator, only a precise observer of what the translator produced}. Empirically, back-translated explanations agree with expert judgments in 99.5\% of cases, compared with at most 94.4\% for direct NL--STL comparison. With this feedback in place, closed-loop accuracy rises to 98.0--99.2\% for strong models. The largest gains appear for weaker models: Qwen2.5-7B-Instruct improves from 45.2\% to 76.0\%, and Qwen2.5-3B-Instruct from 17.6\% to 48.0\%. As in control, the loop also has limits. A forward path too weak to act on corrections cannot be rescued by feedback alone, as with Qwen2.5-0.5B-Instruct, which improves only from 6.8\% to 7.8\%. This result marks when closed-loop refinement is effective.

We evaluate the framework on 500 expert-authored requirements spanning four application domains and a nested-specification subset, using seven LLMs ranging from 0.5B parameters to frontier scale. Beyond accuracy, we examine why feedback helps: replacing semantic feedback with an uninformative retry instruction removes most of the gain. We also examine whether the loop can be trusted, through an expert audit of accepted specifications and a user study comprising 280 interactive sessions.

Our contributions are summarized as follows:
\textbf{(i)} We show that open-loop NL-to-STL translation is somehow ill-posed, because requirements are ambiguous and the written requirement need not match its author's intent. We reformulate the task as closed-loop feedback between human intent and formal specification.
\textbf{(ii)} We develop a closed-loop framework whose feedback path exploits the structure of STL to render generated formulas precisely in natural language. This allows non-experts to validate and correct specifications without reading formal syntax.
\textbf{(iii)} Through experiments on seven LLMs, feedback ablations, an expert audit, and a user study, we show that a precise feedback path substantially reduces dependence on translator capability. We also characterize the capability threshold above which feedback no longer helps.


\section{Related Work}

\textbf{Natural-Language-to-Temporal-Logic Translation.}
Natural-language-to-temporal-logic translation requires preserving
logical structure and temporal semantics across linguistic and
formal representations. Research on LTL and related logics
investigates LLM-based translation~\citep{chen2023nl2tl},
NL-to-LTL tools~\citep{fuggitti2023nl2ltl}, data-efficient
learning~\citep{pan2023data}, and grounding robot instructions
in temporal specifications~\citep{liu2023lang2ltl}.
STL introduces additional requirements for real-valued predicates,
numerical thresholds, and metric temporal intervals.
DeepSTL studies supervised English-to-STL translation using
grammar-generated training data~\citep{he2022deepstl}.
Subsequent methods incorporate LLM-based
generation~\citep{mao2024nl2stl}, external knowledge
augmentation~\citep{fang2025enhancing}, and multi-aspect
reinforcement learning~\citep{fang2026restl}.
ReasonSTL integrates deterministic numerical tools with
structured formula construction to support semantic and
numerical fidelity~\citep{ye2026reasonstl}.
These approaches primarily improve the translator's ability
to recover a formal specification from a supplied requirement.
Our work examines how generated specifications can be assessed
against user intent when the supplied requirement may itself
need revision. We study translation across successive
requirement revisions, treating translator accuracy and
intent alignment as distinct objectives. Advances in the
underlying translation model therefore complement the
protocol studied here.

\textbf{Interactive Requirement Clarification and Specification Refinement.}
A related direction investigates interactive approaches for improving the reliability of language-based formalization. NL2Spec~\cite{cosler2023nl2spec} introduces an interactive framework for translating natural-language descriptions into temporal logic specifications by allowing users to refine intermediate natural-language descriptions associated with logical components. However, the interaction is performed at the level of temporal-logic fragments and requires users to reason about formal structures. In contrast, our setting considers users who provide domain-level requirements without requiring expertise in temporal logic and focuses on validating the semantic meaning of a complete generated specification. More recently, ClarifySTL~\cite{fang2025clarifystl,fang2026clarifystl} studies requirement clarification for STL transformation by identifying ambiguities and acquiring additional information before translation. Such approaches improve the specification input by resolving underspecified requirements. However, input clarification and output validation address different stages of the synthesis pipeline: a well-defined requirement may still be translated into an incorrect STL formula. Our work complements these approaches by introducing feedback after formal specification generation to verify and correct the realized semantics of the generated artifact.

\textbf{Reliability Assessment and Feedback-based Refinement of LLM Outputs.}
Beyond formal specification synthesis, recent studies have investigated improving the reliability of LLM outputs through evaluation and iterative refinement. Self-Refine~\cite{madaan2023self} explores iterative generation with self-feedback, while CRITIC~\cite{gou2024critic} introduces tool-interactive critique for improving model responses. Verification-oriented methods, including process-level supervision and step-wise verification~\cite{lightman2024let}, demonstrate the effectiveness of external evaluation signals for improving reasoning reliability. In addition, SCP-NL2TL~\cite{wang2026scp} investigates selective prediction and semantic verification to identify unreliable temporal-logic translations. However, existing feedback and verification approaches primarily aim to improve output quality or estimate confidence, rather than establishing semantic agreement between a generated formal artifact and the original human intent. In formal specification synthesis, the correctness criterion is not merely whether an output is plausible or verifiable, but whether it preserves the intended meaning of the requirement.

Our work builds upon these directions by introducing a closed-loop semantic acceptance process for NL-to-STL translation. The proposed framework places feedback after specification generation and before downstream formal reasoning, enabling generated specifications to be validated and refined according to their intended semantics. Rather than replacing existing translation or verification techniques, our approach provides an additional alignment layer that connects human intent, generated formal specifications, and subsequent formal analysis.

\section{Preliminaries}
\label{sec:pre}

\subsection{Signal Temporal Logic}
\label{sec:stl}

Signal Temporal Logic (STL) is a formal specification language for real-valued signals in cyber-physical systems. Let $v:[0,T]\rightarrow\mathbb{R}^{d}$ denote a signal trajectory. STL formulas are constructed from real-valued predicates, Boolean connectives, and temporal operators. The syntax is defined recursively as
\begin{equation}
\begin{aligned}
    \phi &::= \pi
    \mid \neg\phi
    \mid \phi_1\wedge\phi_2
    \mid \phi_1\vee\phi_2
    \mid \phi_1\rightarrow\phi_2
    \mid \mathbf{G}_{I}\phi
    \mid \mathbf{F}_{I}\phi
    \mid \phi_1\mathbf{U}_{I}\phi_2,\\
    \pi &::= f(x_1,\ldots,x_k)\sim c ,
\end{aligned}
\label{eq:stl-syntax}
\end{equation}
where $f(x_1,\ldots,x_k)$ is a real-valued signal expression, $\sim\in\{<,\leq,=,\geq,>\}$, $c\in\mathbb{R}$, and $I=[a,b]$ is a temporal interval with $0\leq a\leq b\leq\infty$. The operators $\mathbf{G}_{I}$, $\mathbf{F}_{I}$, and $\mathbf{U}_{I}$ denote the globally, eventually, and until operators, respectively.

The satisfaction relation $(v,t)\models\phi$ defines whether a signal trajectory satisfies an STL formula at time $t$. For a predicate,
\begin{equation}
(v,t)\models f(x_1,\ldots,x_k)\sim c
\end{equation}
if and only if the predicate evaluates to true at time $t$. The temporal operators are interpreted over the associated interval. Specifically,
\begin{equation}
\begin{aligned}
(v,t)\models \mathbf{G}_{[a,b]}\phi
&\iff
\forall t'\in[t+a,t+b],\ (v,t')\models\phi,\\
(v,t)\models \mathbf{F}_{[a,b]}\phi
&\iff
\exists t'\in[t+a,t+b],\ (v,t')\models\phi,\\
(v,t)\models \phi_1\mathbf{U}_{[a,b]}\phi_2
&\iff
\exists t'\in[t+a,t+b],
\end{aligned}
\end{equation}
where $(v,t')\models\phi_2$ and $(v,\tau)\models\phi_1$ for all $\tau\in[t,t']$. STL additionally admits quantitative robustness semantics, which assigns a real-valued measure indicating the degree of satisfaction or violation of a specification.

Throughout this work, we consider the future-time fragment of STL with explicit temporal intervals. All temporal operators are associated with bounded or explicitly specified intervals, with unbounded requirements represented as $[0,\infty)$. STL formulas are represented internally as structured syntax trees for subsequent processing.

\subsection{NL-to-STL Translation}
\label{sec:task}

We consider the task of translating a natural-language requirement into an STL specification. Let $I$ denote the underlying human intent, $x$ denote its natural-language expression, and $\phi$ denote the generated STL formula. An NL-to-STL translator aims to synthesize a specification whose semantics faithfully represents the intended requirement:
\begin{equation}
    \mathrm{sem}(\phi) \equiv I ,
\end{equation}
where $\mathrm{sem}(\phi)$ denotes the behavior characterized by the generated STL formula. This formulation distinguishes three different objects involved in specification synthesis: the latent intent of the requirement, its linguistic description, and the resulting formal representation.

The correctness of NL-to-STL translation is therefore determined by semantic alignment rather than syntactic similarity. Exact matching between a generated formula and a reference specification is insufficient, since semantically equivalent STL formulas may have different syntactic structures. On the other hand, determining semantic equivalence between arbitrary STL formulas is challenging due to the combination of real-valued predicates, temporal intervals, and nested temporal operators. Consequently, verifying whether a generated specification faithfully preserves the intended meaning remains a fundamental challenge.

Moreover, natural-language requirements may themselves be insufficient to determine a unique formal specification. For example, requirements may omit quantitative thresholds, temporal bounds, or triggering conditions required for constructing a complete STL formula. In such cases, additional assumptions are necessary to obtain a specification, and the resulting formula may reflect assumptions introduced by the translator rather than information contained in the original requirement. Therefore, reliable NL-to-STL translation requires distinguishing between ambiguity originating from the requirement and semantic deviations introduced during formalization.
\section{Closed-Loop Semantic Acceptance Framework}
\label{sec:method}

\begin{figure}[t]
    \centering
    \includegraphics[width=0.95\linewidth]{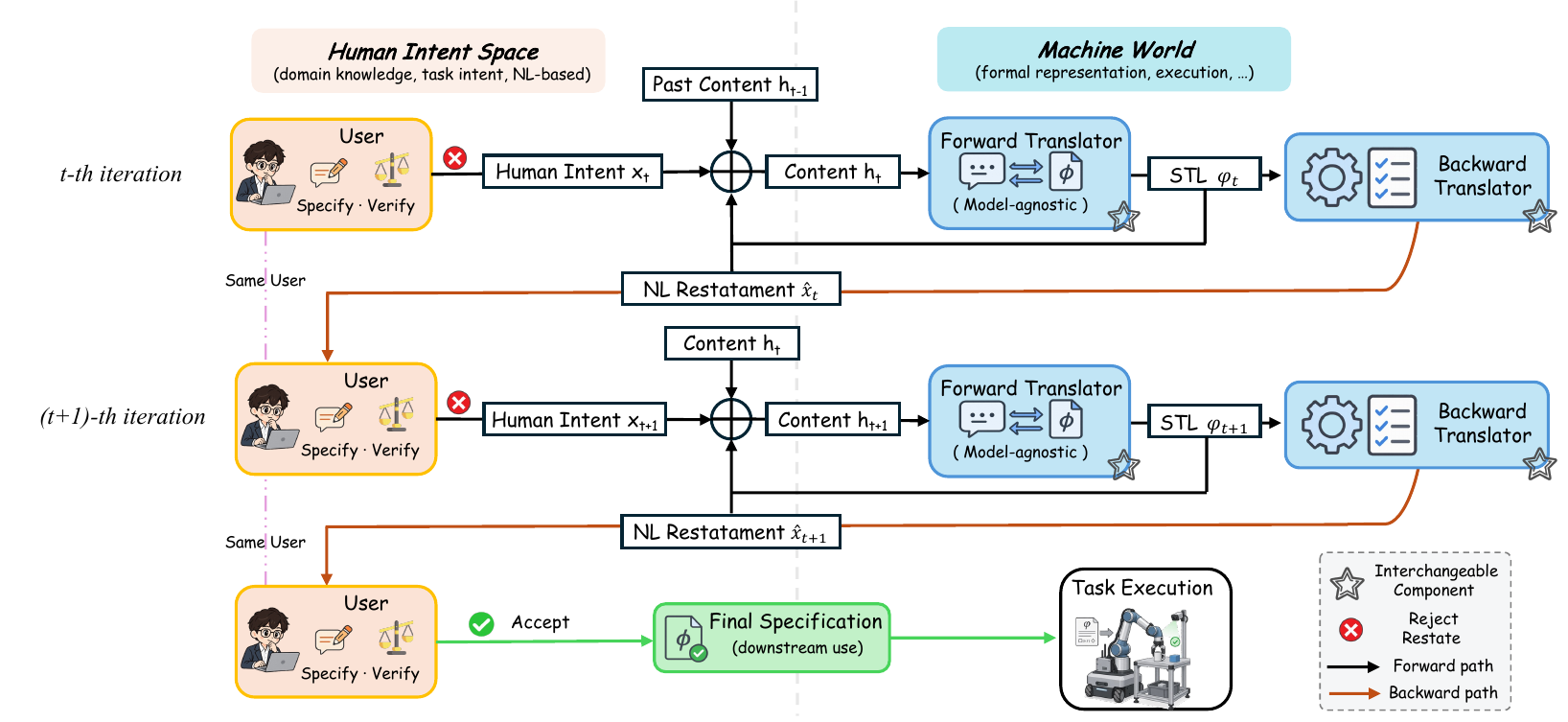}
    \vspace{-0.6em}
    \caption{Closed-loop semantic acceptance for NL-to-STL translation.}
    \vspace{-1.6em}
    \label{fig:framework}
\end{figure}

Natural-language requirements may be incomplete at the outset, while their
translation into STL may introduce additional semantic deviations. We address
both sources of uncertainty through a recurrent interaction between the
framework and a single user. At round $t$, the user supplies a natural-language
utterance $x_t$. The initial requirement is $x_0$; subsequent utterances may
clarify missing details, correct a misinterpretation, or extend the
requirement. Each utterance is incorporated into the interaction context
before a new candidate specification is generated. The resulting specification
is then rendered in natural language and presented to the user for
verification, as illustrated in Figure~\ref{fig:framework}.

\subsection{Iterative Context Update and Semantic Observation}
\label{sec:refinement}

Let $h_t$ denote the compressed interaction context at round $t$.
The framework updates this context using the current user utterance and
the artifacts from the preceding round, translates the updated context
into a candidate STL specification, and renders that specification back
into natural language:
\begin{equation}
\label{eq:closed_loop_update}
\begin{aligned}
h_t &= C(h_{t-1},x_t,\varphi_{t-1},\hat{x}_{t-1}),\\
\varphi_t &= F(h_t), \qquad
\hat{x}_t = B(\varphi_t).
\end{aligned}
\end{equation}
Here, $C$ is the context update operator, $F$ is the forward NL-to-STL
translator, and $B$ is the backward STL-to-NL translator. The preceding
context and artifacts are empty at $t=0$. Consequently, the initial
requirement and all later revisions are processed by the same update rule.

The backward translation $\hat{x}_t$, referred to as \emph{NL feedback},
is a natural-language restatement of the semantics represented by
$\varphi_t$. It is derived from the candidate specification rather than
from the user's presumed intention. To support meaningful verification,
the restatement must reflect the predicates, quantitative parameters,
temporal intervals, and logical relationships expressed in the STL
formula. In particular, formal details such as thresholds and time bounds
should be obtained from $\varphi_t$, rather than inferred independently
during backward translation. The resulting NL feedback provides a
user-facing view of the candidate specification without requiring the
user to inspect STL syntax.

\subsection{User Verification and Requirement Refinement}
\label{sec:acceptance}

The user evaluates whether $\hat{x}_t$ is consistent with their current
intent and provides a decision
$a_t\in\{\mathrm{accept},\mathrm{revise}\}$. If the candidate is accepted,
$\varphi_t$ is returned as the final specification. Otherwise, the user
supplies a new utterance $x_{t+1}$ after reviewing the NL feedback.

A revision may address different sources of mismatch. When the
requirement is underspecified, $x_{t+1}$ can supply information needed
to determine the intended STL semantics, such as a threshold, temporal
bound, or triggering condition. When the requirement is sufficiently
specified but its translation is inaccurate, $x_{t+1}$ can correct the
deviation revealed by $\hat{x}_t$. The user may also introduce an
additional constraint. These cases do not require separate interaction
paths: each new utterance is incorporated into the context through
Eq.~\eqref{eq:closed_loop_update} before the next candidate is generated.

Acceptance is based on the user's assessment of the displayed NL
feedback. Its interpretation therefore depends on the fidelity of the
backward translation; user acceptance alone does not constitute a formal
proof that $\varphi_t$ is equivalent to the intended requirement.

\subsection{Closed-Loop Acceptance Protocol}
\label{sec:protocol}

The protocol alternates between context update, forward translation,
backward translation, and user verification. When the user requests a
revision, the next utterance is informed by the preceding NL feedback,
while the context update retains information from earlier rounds.
This recurrent structure permits both clarification of incomplete
requirements and correction of translation errors within a single
procedure. Algorithm~\ref{alg:closedloop} summarizes the protocol.

\begin{algorithm}[t]
\caption{Closed-loop semantic acceptance for NL-to-STL translation}
\label{alg:closedloop}
\begin{algorithmic}[1]
\REQUIRE Initial user utterance $x_0$, maximum number of rounds $K$
\STATE Initialize
    $h_{-1}\leftarrow\emptyset$,
    $\varphi_{-1}\leftarrow\emptyset$,
    $\hat{x}_{-1}\leftarrow\emptyset$
\FOR{$t=0,\ldots,K-1$}
    \STATE Update context:
        $h_t\leftarrow
        C(h_{t-1},x_t,\varphi_{t-1},\hat{x}_{t-1})$
    \STATE Generate candidate STL:
        $\varphi_t\leftarrow F(h_t)$
    \STATE Generate NL feedback:
        $\hat{x}_t\leftarrow B(\varphi_t)$
    \STATE Present $\hat{x}_t$ and obtain the user's decision
        $a_t\in\{\mathrm{accept},\mathrm{revise}\}$
    \IF{$a_t=\mathrm{accept}$}
        \RETURN $\varphi_t$
    \ENDIF
    \IF{$t<K-1$}
        \STATE Obtain the user's next utterance $x_{t+1}$
    \ENDIF
\ENDFOR
\RETURN $\mathrm{unresolved}$
\end{algorithmic}
\end{algorithm}

The protocol terminates when the user accepts a candidate specification
or when the round budget is exhausted. By placing user verification
between candidate generation and specification delivery, the framework
can resolve missing requirements and semantic deviations before the
formal artifact is used for downstream reasoning. To make this
interaction concrete, we next present two representative execution
traces from our evaluation. The first shows a successful correction
after user feedback, while the second shows an unresolved case after
the full ten-round budget. Both cases use Qwen2.5-3B-Instruct as the
forward translator.

\noindent\textbf{Successful local repair.}
The user utterance is ``The projector blackout blind must remain closed
for the next 5 minutes.'' The first-round candidate is
\(\varphi_0=\mathbf{F}_{[0,300]}\texttt{blind\_closed}\), whose
back-translation reads ``the blind is closed at some point within the
next 300 seconds.'' The gap is visible at the natural-language layer:
the candidate asserts a single instant of closure, while the user
requires closure over the entire window. The user issues \textsc{revise}
with ``the blind must stay closed throughout the next 5 minutes; being
closed at one instant is not enough.'' Conditioned on this feedback, the
translator emits
\(\varphi_1=\mathbf{G}_{[0,300]}\texttt{blind\_closed}\), whose
back-translation confirms invariance over \([0,300]\). The user issues
\textsc{accept}, and the protocol returns \(\varphi_1\) after two rounds.
The feedback localizes a single operator substitution, from the
eventuality operator \(\mathbf{F}\) to the invariance operator
\(\mathbf{G}\), which is sufficient for acceptance.

\noindent\textbf{Unresolved compositional repair.}
The second utterance couples four ordered constraints: the robot stays
stationary in a room during the first 15 minutes after 7:00 AM, picks up
food between 7:15 and 7:20 AM, delivers it within 2 minutes of pickup,
and holds speed below 2 m/s throughout the operation. Normalizing
7:00 AM to \(t=0\), the intended semantics is
\begin{align*}
\varphi^\star={}&
\mathbf{G}_{[0,900]}(\texttt{in\_room}\land\texttt{speed}=0)
\land \mathbf{F}_{[900,1200]}e_{\mathrm{pick}} \\
&{}\land
\mathbf{G}_{[900,1200]}
\bigl(e_{\mathrm{pick}}\rightarrow
\mathbf{F}_{[0,120]}\texttt{delivered}\bigr)
\land \mathbf{G}_{[0,1320]}(\texttt{speed}<2),
\end{align*}
where \(e_{\mathrm{pick}}\) is a one-time pickup event and the final
bound spans the longest admissible operation, with pickup at
\(t=1200\) followed by the 2-minute delivery deadline.

The first-round candidate is
\(\varphi_0=\mathbf{F}_{[0,15]}(\texttt{robot\_moving}=0)\). It exhibits
three independent errors: an eventuality operator in place of an
invariance operator, a unit error that reads 15 minutes as 15 seconds,
and the omission of the pickup, delivery, and speed constraints. The
user issues \textsc{revise} and restates all four requirements, with the
continuous stationarity window, the pickup deadline, the pickup-triggered
delivery bound, and the operation-wide speed limit. By round 4 the
translator repairs the first constraint and emits
\(\mathbf{G}_{[0,900]}\texttt{stationary\_in\_room}\), whose
back-translation reports stationarity over the first 900 seconds. The
remaining three constraints stay absent. Each subsequent user decision
is \textsc{revise} and re-specifies the same missing structure, yet
rounds 4 through 10 reproduce the identical candidate. The tenth-round
formula still contains no pickup event, no triggered delivery deadline,
and no speed bound, so the protocol returns \(\mathrm{unresolved}\) once
the budget \(K=10\) is spent. The trace isolates a compositional repair
failure: back-translation exposes every missing conjunct and the user
supplies targeted feedback, but the translator converges to a single
corrected clause without recovering the full temporal composition.

\section{Experiments}
\label{sec:experiments}

We evaluate the proposed closed-loop semantic acceptance framework from three aspects:
(1) its effectiveness in improving NL-to-STL translation over one-shot generation;
(2) the impact of different feedback components and model capabilities on iterative refinement; and
(3) the reliability of the acceptance loop through human evaluation and robustness analysis.

\subsection{Experimental Setup}
\label{sec:setup}

\textbf{Benchmark and Models.}
We evaluate all models on a benchmark comprising 500 expert-authored natural-language requirements from four application domains: autonomous driving, AGVs, drones, and smart homes, together with a dedicated subset of nested temporal specifications. The benchmark was constructed by five domain experts and contains 335 well-specified requirements, 122 STL translation error cases, and 43 underspecified requirements. The 500 requirements were randomly sampled from the collected instances using a fixed random seed. Details of the benchmark construction process and annotation protocol are provided in the Appendix.

We evaluate seven LLMs spanning a broad range of model scales:
Qwen2.5-0.5B-Instruct, Qwen2.5-3B-Instruct, Qwen2.5-7B-Instruct, Qwen3.8-27B, Qwen3.6-Flash, Qwen3.7-Max, and Qwen3.8-Max. The open-loop setting performs one-shot NL-to-STL translation without refinement, whereas the closed-loop setting enables iterative semantic refinement with a maximum interaction budget of 10 rounds.

\textbf{Evaluation Protocol.}
Except in experiments explicitly designated as human-in-the-loop, closed-loop evaluation uses an LLM-as-Judge to provide semantic acceptance decisions and, upon rejection, simulated follow-up NL inputs. A feedback event is a noninitial NL input $x_t$ ($t\geq 1$) that clarifies or corrects the requirement; thus, a run accepted in the first round has zero feedback events. We report open-loop semantic accuracy, closed-loop convergence rate, feedback frequency, interaction cost, and performance across refinement budgets. A run converges only if its STL specification is accepted within the prescribed budget. The acceptance mechanism is assessed through expert review and controlled perturbation experiments, with implementation details in Appendix.
\subsection{Semantic Feedback Improves Specification Fidelity}
\label{sec:main}

We compare one-shot NL-to-STL translation with closed-loop refinement
to assess changes in semantic accuracy and interaction cost.
Table~\ref{tab:main_results} reports the results across seven models.
For the benchmark-scale evaluation, an automated semantic acceptance
module supplies acceptance decisions and follow-up NL inputs. Its
agreement with human assessments is examined in
Section~\ref{sec:reliability}.

Closed-loop accuracy exceeds open-loop accuracy for every evaluated
model, with gains ranging from 1.0 to 30.8 percentage points. The
largest gains are observed for Qwen2.5-3B-Instruct and
Qwen2.5-7B-Instruct, which improve by 30.4 and 30.8 percentage points,
respectively. These gains require an average of 5.34 and 2.83
additional NL inputs per sample, indicating that the accuracy
improvement for these models entails substantial interaction.

The response to feedback is strongly model-dependent.
Qwen2.5-0.5B-Instruct reaches only 7.8\% closed-loop accuracy, an
increase of 1.0 percentage point despite 8.34 feedback events per
sample. This pattern is consistent with limited capacity to
incorporate corrective information and shows that additional
refinement rounds do not uniformly yield meaningful gains.

For the four higher-performing models, closed-loop accuracy reaches
98.0--99.2\%, while mean feedback remains below 0.5 events per
sample. Their improvements range from 10.2 to 17.6 percentage points
despite open-loop accuracies above 80\%. Qwen3.7-Max has the highest
open-loop accuracy (89.0\%) and reaches 99.2\% after refinement.
Qwen3.8-Max reaches the same closed-loop accuracy with the lowest
mean feedback count (0.19 per sample). These results characterize
the measured gains and their interaction costs under the automated
acceptance protocol; Section~\ref{sec:reliability} evaluates the
reliability of the acceptance decisions on which the protocol depends.

\begin{table}[t]
\centering
\vspace{-1mm}
\caption{\small
Comparison between open-loop translation and closed-loop semantic refinement.
Gain is reported in percentage points (pp).}
\label{tab:main_results}
\scalebox{0.82}{
\begin{tabular}{lcccc}
\toprule
Model
& Open-loop Acc$\uparrow$
& Closed-loop Acc$\uparrow$
& Gain (pp)$\uparrow$
& Feedback/sample$\downarrow$ \\
\midrule
Qwen2.5-0.5B-Instruct & 6.8\%  & 7.8\%  & +1.0  & 8.34 \\
Qwen2.5-3B-Instruct   & 17.6\% & 48.0\% & +30.4 & 5.34 \\
Qwen2.5-7B-Instruct   & 45.2\% & 76.0\% & +30.8 & 2.83 \\
Qwen3.8-27B           & 86.4\% & 98.0\% & +11.6 & 0.41 \\
Qwen3.6-Flash         & 81.4\% & 99.0\% & +17.6 & 0.32 \\
Qwen3.7-Max           & 89.0\% & 99.2\% & +10.2 & 0.20 \\
Qwen3.8-Max           & 88.6\% & 99.2\% & +10.6 & 0.19 \\
\bottomrule
\end{tabular}}
\vspace{-1.78em}
\end{table}

\subsection{Understanding the Role of Semantic Feedback}
\label{sec:feedback_analysis}

We examine whether closed-loop gains arise from the semantic content of
feedback rather than repeated generation, and how models use feedback
across successive refinement rounds.

\begin{wrapfigure}{r}{0.50\columnwidth}
\centering
\vspace{-1.6em}
\scalebox{0.82}{\textbf{(a) Feedback ablation}}\par
\vspace{0.4em}

\begingroup
\setlength{\tabcolsep}{2pt}
\scalebox{0.82}{%
\begin{tabular*}{1.22\linewidth}{@{\extracolsep{\fill}}lccc@{}}
\toprule
Model & I & U & $\Delta$ \\
\midrule
Qwen2.5-3B  & 44.0\% & 17.5\% & +26.5 \\
Qwen2.5-7B  & 76.0\% & 50.0\% & +26.0 \\
Qwen3.8-27B & 98.0\% & 88.0\% & +10.0 \\
Qwen3.8-Max & 99.5\% & 89.9\% & +9.6  \\
\bottomrule
\end{tabular*}}
\endgroup

\vspace{0.7em}
\scalebox{0.82}{\textbf{(b) Interaction dynamics}}\par
\vspace{0.3em}
\includegraphics[width=\linewidth]{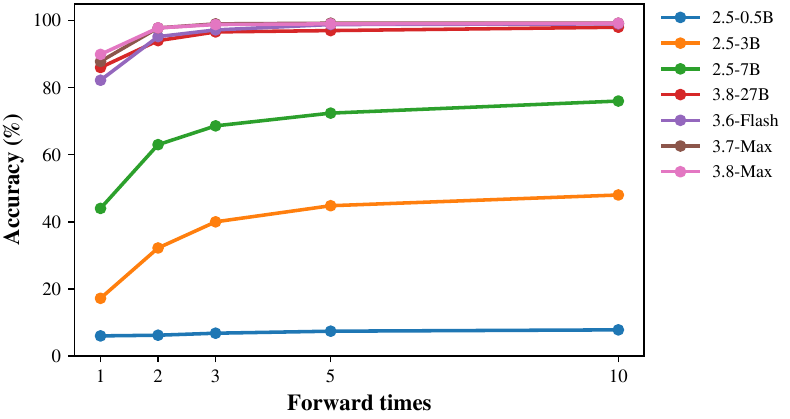}

\caption{\small
Feedback ablation and convergence.
I/U: informative/uninformative feedback;
$\Delta$: accuracy difference in percentage points. All models are Qwen models.
}
\vspace{-0.4em}
\label{fig:feedback_analysis}
\end{wrapfigure}

To assess whether the gains arise from semantic guidance rather than
repeated generation, we replace each corrective input with a generic
retry instruction. This instruction tells the model that its previous
specification is incorrect and asks it to try again, but does not
identify the erroneous predicate, temporal operator, interval, or
omitted constraint. The model and refinement procedure remain
unchanged, so both conditions allow further translation attempts;
only the informative condition provides a specific direction for
correction. Figure~\ref{fig:feedback_analysis}(a) shows that
informative feedback yields higher accuracy for all four evaluated
models, with differences of 9.6--26.5 percentage points. The largest
differences occur for Qwen2.5-3B-Instruct and Qwen2.5-7B-Instruct,
at 26.5 and 26.0 percentage points, respectively. These results
support the conclusion that the semantic content of feedback
contributes to refinement beyond a rejection signal and another
generation opportunity.

Figure~\ref{fig:feedback_analysis}(b) examines the distribution of
gains across interaction rounds. Higher-capability models approach
their final performance within the first few rounds, whereas weaker
models benefit less from continued refinement. In particular,
Qwen2.5-0.5B-Instruct improves from 6.8\% open-loop accuracy to only
7.8\% closed-loop accuracy, despite access to further attempts.
The capacity to interpret and incorporate a correction therefore
appears to constrain the benefit obtainable from feedback.

We also measure interaction cost as the total number of translation
rounds across all samples divided by the number of accepted
specifications. This aggregate ratio is approximately 1.2 for the
highest-performing models, compared with 119.7 for
Qwen2.5-0.5B-Instruct. It can exceed the ten-round per-session budget
because rounds spent on unsuccessful sessions contribute to the
numerator. Together, the ablation and interaction analyses indicate
that semantic feedback supplies useful corrective information, while
translator capability determines how efficiently that information
can be converted into an accepted specification.

\subsection{Reliability of the Semantic Acceptance Loop}
\label{sec:reliability}


\begin{table*}[t]
\centering
\caption{\small
Automated acceptance reliability and real-user evaluation.
$D_1$: direct NL--STL with DeepSeek-v4-Flash;
$D_2$: direct NL--STL with Qwen3.6-27B;
$B_3$: three-model back-translation.}
\label{tab:acceptance_and_human}

\begin{minipage}[t]{0.49\textwidth}
\vspace{0pt}
\raggedright
\scalebox{0.82}{
\begin{tabular}{llc}
\toprule
\multicolumn{3}{l}{\textbf{(a) Acceptance reliability}} \\
\midrule
Evaluation & Method or metric & Result \\
\midrule
Calibration & $D_1$ & 92.8\% \\
            & $D_2$ & 94.4\% \\
            & $B_3$ & \textbf{398/400 (99.5\%)} \\
\midrule
Human audit & Acceptance precision & 629/634 (99.2\%) \\
            & Wilson 95\% CI      & [98.2\%, 99.7\%] \\
\bottomrule
\end{tabular}}
\end{minipage}\hfill
\begin{minipage}[t]{0.47\textwidth}
\vspace{0pt}
\raggedright
\scalebox{0.82}{
\begin{tabular}{lcc}
\toprule
\multicolumn{3}{l}{\textbf{(b) Real-user evaluation}} \\
\midrule
Model & Acceptance rate$\uparrow$ & FB/sample$\downarrow$ \\
\midrule
Qwen2.5-0.5B-Instruct & 7.5\%   & 7.43 \\
Qwen2.5-3B-Instruct   & 50.0\%  & 4.55 \\
Qwen2.5-7B-Instruct   & 62.5\%  & 3.73 \\
Qwen3.8-27B           & 100.0\% & 0.30 \\
Qwen3.6-Flash         & 95.0\%  & 0.70 \\
Qwen3.7-Max           & 92.5\%  & 1.20 \\
Qwen3.8-Max           & 97.5\%  & 0.35 \\
\bottomrule
\end{tabular}}
\end{minipage}
\vspace{-1.6em}
\end{table*}

We assess the reliability of automated semantic acceptance through
interface-level calibration and a separate human audit. As shown in
Table~\ref{tab:acceptance_and_human}(a), direct NL--STL comparison
achieves 92.8\% and 94.4\% agreement with expert labels on 400 cases.
The back-translation interface achieves 398/400 agreement (99.5\%),
suggesting that an explicit NL representation of the candidate STL
provides a more reliable basis for acceptance decisions in this
calibration. We further audit 634 specifications accepted by the
complete framework. Experts confirm 629 as semantically correct,
yielding an acceptance precision of 99.2\% (Wilson 95\% CI:
[98.2\%, 99.7\%]); the remaining five accepted specifications
(0.8\%) are judged incorrect.

We also evaluate the framework with real users on 40 requirements
shared across seven models, yielding 280 model--requirement sessions.
In each session, participants inspect the NL restatement of a candidate
STL specification and either accept it or submit a revised NL input.
Within the ten-round budget, 202 sessions (72.1\%) reach participant
acceptance. Participants provide 1,010 NL inputs in total: 280 initial
requirements and 730 subsequent feedback inputs that lead to another
translation attempt. As shown in
Table~\ref{tab:acceptance_and_human}(b), acceptance among the locally
deployed models increases from 7.5\% for Qwen2.5-0.5B-Instruct to
100.0\% for Qwen3.8-27B, while mean feedback per sample decreases
from 7.43 to 0.30. The three models with larger parameter counts
(Qwen3.6-Flash, Qwen3.7-Max, and Qwen3.8-Max) achieve user acceptance
rates of 92.5--97.5\% while requiring 0.35--1.20 additional NL inputs
per sample. Thus, these models attain high acceptance within the
ten-round budget with relatively few feedback inputs.

\section{Conclusion}

This work revisits NL-to-STL translation from the perspective of semantic alignment rather than one-shot generation. While formal methods provide rigorous guarantees for systems satisfying a specification, they do not ensure that the specification itself faithfully represents human intent. By introducing a closed-loop feedback mechanism before downstream formal reasoning, we enable generated specifications to be validated and refined through natural-language interaction. Our results demonstrate that feedback can substantially improve specification fidelity while revealing the importance of feedback quality and model capability. More broadly, this work establishes semantic feedback as an alignment layer between human intent and machine-executable formal artifacts. Formal verification guarantees correctness with respect to a given specification; semantic feedback ensures that the specification being verified is the one intended by its human author.

\newpage
\subsection*{Acknowledgments}

We thank the four human experts who contributed to the construction of the
dataset by providing domain knowledge, annotating data samples, and reviewing
the quality of the collected data. Their valuable efforts and insights were
essential for building a reliable and meaningful benchmark for this work.

\bibliography{iclr2027_conference}
\bibliographystyle{iclr2027_conference}
\newpage

\appendix


\section{Benchmark Construction}
\label{app:benchmark}

\noindent\textbf{Data collection.} The benchmark covers advanced
driver-assistance systems (ADAS), unmanned aerial vehicles (Drone),
smart homes, and automated guided vehicles (AGV). Five domain experts
wrote the requirements. Two experts independently covered AGV because
it spans a broader set of navigation, scheduling, and safety tasks.
All annotators followed the same STL guideline and passed a
qualification exercise.

We retained 3,556 of the initial 4,000 requirements after consistency
checks and expert adjudication. Table~\ref{tab:dataset_statistics}(a)
reports the resulting domain distribution. The larger AGV share follows
the planned allocation of two independent annotators to that domain.
The distribution was fixed before sampling, and the other three domains
remain within 5.2 percentage points of one another.

\begin{table}[!b]
\centering
\vspace{-1mm}
\caption{\small
Statistics of the benchmark and evaluation set.
}
\label{tab:dataset_statistics}

\begin{minipage}{0.48\linewidth}
\centering
\caption*{\small (a) Domain distribution}
\scalebox{0.90}{
\begin{tabular}{lcc}
\toprule
Domain & Instances & Proportion \\
\midrule
ADAS       & 669   & 18.8\% \\
Drone      & 724   & 20.4\% \\
Smart Home & 853   & 24.0\% \\
AGV        & 1,310 & 36.8\% \\
\midrule
Total      & 3,556 & 100.0\% \\
\bottomrule
\end{tabular}}
\end{minipage}
\hfill
\begin{minipage}{0.48\linewidth}
\centering
\caption*{\small (b) Evaluation-set composition}
\scalebox{0.90}{
\begin{tabular}{lcc}
\toprule
Instance type & Count & Proportion \\
\midrule
Well-specified       & 335 & 67.0\% \\
Translation error    & 122 & 24.4\% \\
Underspecified input & 43  & 8.6\% \\
\midrule
Total                & 500 & 100.0\% \\
\bottomrule
\end{tabular}}
\end{minipage}

\end{table}

The retained set contains 1,455 simple, 1,640 medium, and 461 hard
requirements. Among them, 1,100 are annotated as formula errors or
underspecified requirements, including 402 minimal contrastive pairs in
which the requirement is held fixed and the candidate formulas differ
in one semantic dimension. Consequently, the benchmark evaluates formula
generation and whether a system can distinguish an output-side
mistranslation from information
that is absent at the input.

\noindent\textbf{Comparison with existing corpora.}
Table~\ref{tab:corpus-comparison} positions the benchmark relative to
three representative NL-to-STL corpora. The comparison is based on the
supervision released or explicitly described by each work; ``--'' means
that a dimension is not reported as an annotation target.

\begin{table*}[t]
\centering
\vspace{-1mm}
\caption{\small
Comparison with representative NL-to-STL corpora. Dataset size alone is
not a quality measure; the lower rows distinguish translation corpora
from data that explicitly supervise semantic diagnosis and correction.
}
\label{tab:corpus-comparison}
\resizebox{\textwidth}{!}{
\begin{tabular}{lcccc}
\toprule
Property
& DeepSTL~\cite{he2022deepstl}
& STL-DivEn~\cite{fang2025enhancing}
& \textsc{STL-Bench}~\cite{ye2026reasonstl}
& \textbf{Ours} \\
\midrule
NL--STL pairs & $\sim$120K & $\sim$16K & 28,880 & 3,556 \\
Language & English & English & English + Chinese & English \\
Construction
& Grammar-generated
& Seed + LLM augmentation
& Scenario/template + LLM
& \textbf{Expert-authored} \\
Domain grounding
& Symbolic
& Partial
& 6 domains / 33 scenarios
& 4 CPS domains \\
Formula-level error labels & -- & -- & -- & \textbf{759} \\
Underspecified-input labels & -- & -- & -- & \textbf{341} \\
Minimal contrastive pairs & -- & -- & -- & \textbf{402} \\
Error explanation and correction & -- & -- & -- & \textbf{Yes} \\
Whole-corpus expert cross-review & -- & -- & -- & \textbf{Yes} \\
\bottomrule
\end{tabular}}
\end{table*}

The comparison makes the intended trade-off explicit. Existing corpora
provide substantially more correct translation pairs and are therefore
valuable for large-scale training. Our benchmark serves a different
purpose. Domain experts write and cross-review its requirements and
defect annotations, which limits scale while supporting diagnostic
supervision: formula-level errors,
missing input information, localized explanations, corrected targets,
and controlled minimal pairs are represented in a single corpus. These
annotations provide the signals used here to evaluate the two feedback
paths in the closed loop.

\noindent\textbf{Annotation and quality control.} Each instance pairs a
natural-language requirement with an STL specification and semantic
metadata. Well-specified cases admit a complete formula.
Translation-error cases pair a sufficient requirement with an incorrect
candidate and its correction. Underspecified cases omit a threshold,
temporal bound, trigger, scope condition, or other information needed
for formalization.

Experts first judged whether the requirement was sufficient and then
checked the STL, keeping input uncertainty separate from translation
error. Cross-review raised 123 substantive objections. Adjudication
revised and retained 120 cases and removed three for which agreement was
not reached. These revisions show that cross-review materially affected
the released annotations.

\noindent\textbf{Nested temporal subset.} The 81-case Nested subset
covers temporal anchoring, operator scope, event--state distinctions,
nested implication, and dependencies across temporal windows. These
cases are designed to test structural alignment beyond lexical overlap.

\noindent\textbf{Main evaluation set.} Every translator and both
evaluation conditions use the same random sample of 500 cases:
335 well-specified requirements, 122 translation errors, and 43
underspecified inputs (Table~\ref{tab:dataset_statistics}(b)). Thus,
one third of the set requires either output correction or input
clarification. This composition deliberately stresses both feedback paths;
it should not be interpreted as an estimate of their prevalence in
deployed requirements.

\section{System Implementation and Component Validation}
\label{app:implementation}

\noindent\textbf{Closed-loop protocol.}
Following the notation in Section~\ref{sec:method}, the forward
translator \(F\) generates a candidate specification
\(\varphi_t=F(h_t)\), and the backward translator \(B\) renders the
semantic observation \(\hat{x}_t=B(\varphi_t)\). At the implementation
level, an acceptance mechanism compares the requirement with
\(\hat{x}_t\) and returns a decision
\(a_t\in\{\mathrm{accept},\mathrm{revise}\}\) together with a
localized discrepancy \(\delta_t\) upon rejection. This automated
mechanism instantiates the user-verification step in the benchmark-scale
evaluation; human participants make the decision in the real-user study.
The system converts \(\delta_t\) into a natural-language correction that
is appended to the interaction context \(h_t\) for the next round. The
budget is \(K=10\) rounds. A session ends upon acceptance, budget exhaustion, or a
system-side failure. Algorithm~\ref{alg:closedloop} gives the complete
procedure; this section reports the implementation and evaluation
details omitted there.

The reference STL is withheld from \(F\), \(B\), the acceptance
mechanism, and the simulated user; it is used only for external
assessment. For an underspecified case, the simulated user and automated
acceptance mechanism receive the expert-authored corrected NL as an
intent anchor. The forward translator does not receive this anchor
directly and sees only the clarification produced through the simulated
interaction. This design uses the corrected NL to represent latent user
intent while preventing direct leakage of the reference formula.

\noindent\textbf{Translator.} The translator receives the requirement,
available signal definitions, and accumulated feedback. Its prompt
requires exact preservation of predicates, polarity, thresholds,
units, Boolean dependencies, temporal operators, and interval
boundaries. The full feedback history is retained across rounds.

For underspecified inputs, the system requests the smallest clarification
needed to make formalization possible and does not ask the translator to
guess a missing value. For translation errors,
feedback identifies the semantic discrepancy without revealing the
reference STL.

\noindent\textbf{Back-translation.} The backward translator \(B\) parses
\(\varphi_t\) into an abstract syntax tree and renders an inspectable account of its
semantics. The rendering preserves:

\begin{itemize}
    \item predicates, variables, comparison directions, and thresholds;
    \item negation and Boolean dependencies;
    \item temporal operators and interval boundaries;
    \item implication antecedents and consequents;
    \item the scope and nesting of temporal subformulas.
\end{itemize}

The input to \(B\) is restricted to \(\varphi_t\); it cannot inspect
\(x\), the reference formula, or earlier candidates. Thus, the rendered
text describes the candidate without attempting to repair it. This
converts cross-representation comparison into natural-language alignment
while preserving the structure of the formal artifact.

\noindent\textbf{Acceptance mechanism.} In automated evaluation, the
mechanism compares the current intent anchor with \(\hat{x}_t\); in the
real-user study, the participant performs this comparison. Acceptance
requires agreement in predicates, polarity, thresholds, units,
temporal bounds, logical dependencies, and temporal scope. A
candidate is rejected when any required condition is missing,
contradicted, or assigned to the wrong temporal context.

Rejections follow one of two paths. Missing information in \(x\)
triggers clarification; a mismatch between a sufficient \(x\) and
\(\hat{x}_t\) triggers corrective feedback. The first path avoids invented
requirements. The second assigns a detected translation error to the
translator and does not treat it as missing user information.

\noindent\textbf{Prompt interfaces.} The operational prompts enforce
the component-specific constraints
summarized in Table~\ref{tab:prompt-responsibilities}.

\begin{table*}[t]
\centering
\vspace{-1mm}
\caption{\small
Prompt responsibilities in the closed-loop system.
}
\label{tab:prompt-responsibilities}

\renewcommand{\arraystretch}{1.25}
\scalebox{0.92}{
\begin{tabular}{p{2.0cm}p{3.2cm}p{6.4cm}}
\toprule
Component & Input & Required behavior \\
\midrule

Sufficiency gate
& Requirement and interaction history
& Identify information that is necessary for formalization but absent
from the current requirement. \\
\midrule

Clarification
& Missing information
& Ask one minimal and answerable question without suggesting an
arbitrary value. \\
\midrule

Translator
& Requirement, signal definitions, and feedback history
& Produce a syntactically valid STL formula while preserving all
confirmed semantic constraints. \\
\midrule

Back-translation
& Generated STL only
& Render the complete formula semantics without consulting the original
requirement or introducing new information. \\
\midrule

Acceptance
& Requirement and semantic observation
& Accept only under semantic agreement; otherwise return a localized
discrepancy. \\
\midrule

Corrective feedback
& Detected discrepancy
& Express the smallest actionable correction without exposing the
reference STL. \\

\bottomrule
\end{tabular}}
\end{table*}

The components in Table~\ref{tab:prompt-responsibilities} serve distinct
functions. Restricting back-translation to the generated
formula prevents it from silently repairing the candidate, while keeping
the reference annotation outside all operational prompts prevents target
leakage. This design requires the loop to localize and communicate a
semantic discrepancy while keeping the gold specification hidden.

A condensed translator instruction is:

\begin{quote}
\small
Given the requirement, signal definitions, and feedback history,
produce one syntactically valid STL formula. Preserve every stated
condition, polarity, threshold, unit, Boolean dependency, temporal
operator, and interval. Use only information provided by the
requirement or confirmed through feedback. Return the STL formula
without additional explanation.
\end{quote}

The back-translation instruction is:

\begin{quote}
\small
Describe the semantics of the given STL formula in natural language.
Preserve every predicate, comparison, threshold, interval, Boolean
dependency, implication, and temporal scope. Read only the STL
formula. Do not infer missing information, repair the formula, or
consult the original requirement.
\end{quote}

The acceptance instruction is:

\begin{quote}
\small
Compare the requirement with the semantic observation. Accept only
if all required predicates, values, polarities, dependencies,
temporal bounds, and scopes agree. If the requirement lacks necessary
information, request clarification. Otherwise, identify the minimal
semantic discrepancy that must be corrected.
\end{quote}

\noindent\textbf{Evaluation metrics.}
Let \(I_i\) denote the intended semantics, \(\varphi_{i,0}\) the first
candidate, and \(\hat{\varphi}_i\) the delivered formula for case \(i\).
Open-loop accuracy is
\(\mathrm{Acc}_{\mathrm{open}}=N^{-1}\sum_{i=1}^{N}
\mathbb{I}[\mathrm{Match}(I_i,\varphi_{i,0})]\). Closed-loop accuracy uses
the operational stopping criterion,
\(\mathrm{Acc}_{\mathrm{closed}}=N^{-1}\sum_{i=1}^{N}
\mathbb{I}[\exists t<K:a_{i,t}=\mathrm{accept}]\). In the acceptance-noise
ablation, we also report externally judged accuracy,
\(\mathrm{Acc}_{\mathrm{true}}=N^{-1}\sum_{i=1}^{N}
\mathbb{I}[\mathrm{Match}(I_i,\hat{\varphi}_i)]\). If \(L_i\) is the
number of translator calls, interaction cost is
\(\mathrm{Feedback/Case}=N^{-1}\sum_{i=1}^{N}(L_i-1)\). Failed cases
remain in
both the accuracy denominator and the cost calculation. We retain the
full sequence of candidates, observations, decisions, feedback, and
termination states for case-level auditing.

\noindent\textbf{Acceptance-interface calibration.} The 99.5\% figure reports
the agreement of the complete acceptance interface with expert
accept/reject labels. It measures decision quality and does not describe
how fluently \(\hat{x}_t\) reads.
After \(B\) renders the STL abstract syntax tree, DeepSeek-v4-Flash,
Kimi-K3, and GLM-5.3 compare \(x\) with \(\hat{x}_t\); majority vote gives
the final decision. The calibration set contains
400 expert-labeled cases: 249 semantic matches and 151 semantic
mismatches. The interface correctly identifies 248 of 249 matches
and 150 of 151 mismatches. Its overall accuracy is therefore
398/400, or 99.5\%. The mechanical rule layer resolves 92 cases without
an API call.

Two interface errors account for the gap from perfect calibration. The
panel falsely rejects one case whose numerical expression is
semantically equivalent
after unit normalization. It falsely accepts another case where an
unresolved conjunction-disjunction scope changes the intended
semantics. The panel makes no error on the ADAS, Drone, Smart Home,
and Nested cases in the calibration set; both errors fall in the AGV
subset, where accuracy is 97.5\%.

Table~\ref{tab:component-validation} compares this interface with two
direct NL--STL judges. The direct judges are evaluated on 500 cases. The
back-translation interface uses a separate 400-case calibration set, so
the results are not paired. The Nested
column reports recall on the available positive Nested cases and
makes the structural difference between the interfaces explicit.

\begin{table*}[!b]
\centering
\vspace{-1mm}

\begin{minipage}{0.62\linewidth}
\centering
\caption{\small
Validation of semantic comparison interfaces. Overall and Nested
results retain their original denominators because the protocols use
separate evaluation sets.
}
\label{tab:component-validation}
\scalebox{0.88}{
\begin{tabular}{lcc}
\toprule
Interface & Overall accuracy & Nested recall \\
\midrule
Direct: DeepSeek-v4-Flash & 464/500 (92.8\%) & 14/42 (33.3\%) \\
Direct: Qwen3.6-27B       & 472/500 (94.4\%) & 34/44 (77.3\%) \\
Back-translation panel    & \textbf{398/400 (99.5\%)} &
\textbf{81/81 (100.0\%)} \\
\bottomrule
\end{tabular}}
\end{minipage}
\hfill
\begin{minipage}{0.34\linewidth}
\centering
\caption{\small
Closed-loop results on the 81-instance Nested subset.
}
\label{tab:nested-results}
\scalebox{0.90}{
\begin{tabular}{lcc}
\toprule
Model & Correct & Accuracy \\
\midrule
Qwen3.8-27B   & 74/81 & 91.4\% \\
Qwen3.6-Flash & 78/81 & 96.3\% \\
Qwen3.7-Max   & 80/81 & 98.8\% \\
Qwen3.8-Max   & 77/81 & 95.1\% \\
\bottomrule
\end{tabular}}
\end{minipage}

\end{table*}

The two direct judges obtain 33.3\% and 77.3\% Nested recall on their
500-case evaluation. The back-translation interface obtains 81/81 recall
on the positive Nested cases in its separate calibration set. This
result is consistent with the intended role of structured rendering,
which exposes temporal scope, anchoring, and antecedent--consequent
relations before comparison. Different evaluation sets and panel
aggregation prevent a controlled attribution of the observed gap to
structured rendering alone.

\section{Domain-wise Results and Nested Analysis}
\label{app:additional-results}

\noindent\textbf{Complete translation results.} Table~\ref{tab:complete-main-results} reports all seven translators on
the shared 500-instance set under open-loop and closed-loop evaluation.

\begin{table*}[t]
\centering
\vspace{-1mm}
\caption{\small
Open-loop and closed-loop results on the shared 500-instance
evaluation set (reproduced from Table~\ref{tab:main_results} for self-contained reference).
Gain is reported in percentage points (pp).
}
\label{tab:complete-main-results}
\scalebox{0.90}{
\begin{tabular}{lcccc}
\toprule
Model
& Open-loop Acc$\uparrow$
& Closed-loop Acc$\uparrow$
& Gain (pp)$\uparrow$
& Feedback/sample$\downarrow$
\\
\midrule
Qwen2.5-0.5B-Instruct & 6.8\%  & 7.8\%  & +1.0  & 8.34 \\
Qwen2.5-3B-Instruct   & 17.6\% & 48.0\% & +30.4 & 5.34 \\
Qwen2.5-7B-Instruct   & 45.2\% & 76.0\% & +30.8 & 2.83 \\
Qwen3.8-27B           & 86.4\% & 98.0\% & +11.6 & 0.41 \\
Qwen3.6-Flash         & 81.4\% & 99.0\% & +17.6 & 0.32 \\
Qwen3.7-Max           & 89.0\% & 99.2\% & +10.2 & 0.20 \\
Qwen3.8-Max           & 88.6\% & 99.2\% & +10.6 & 0.19 \\
\bottomrule
\end{tabular}}
\end{table*}

The measured interaction gain is non-monotonic across the seven models.
Qwen2.5-0.5B improves by 1.0 percentage point, indicating little benefit
from feedback under this protocol. The 3B and 7B models yield the largest
gains, at 30.4 and 30.8 points. The four higher-performing models begin
with stronger open-loop results, reach 98.0--99.2\% closed-loop accuracy,
and use 0.19--0.41 feedback messages per case. Within this evaluation,
the largest gains occur for the 3B and 7B models. The 0.5B model shows
little net gain despite frequent feedback, while the four
higher-performing models present fewer initial errors.

The observed relation between closed-loop accuracy and feedback cost is also
operationally relevant. Higher-capacity translators reach their reported
accuracy with fewer repair rounds. Within the evaluated set, all four
larger models receive an operational acceptance on every ADAS and AGV
case. Their remaining non-Nested unresolved sessions occur in Drone and
Smart Home, while Nested cases are the only unresolved category shared by
all four. This distribution suggests that structural composition
contributes to the remaining difficulty.

\noindent\textbf{Nested temporal reasoning.} Table~\ref{tab:nested-results}
reports closed-loop performance on the 81-instance Nested subset. Relative
to the corresponding 500-case results, accuracy decreases by 0.4--6.6
percentage points, with the largest reduction observed for Qwen3.8-27B.
The best result in this table is 80 of 81 cases, indicating that most
cases in this subset are resolved under the localized-feedback protocol.

Residual failures mainly involve incorrect temporal anchoring, operator
scope, event--state confusion, and dependencies between nested temporal
windows. These errors are difficult to detect through lexical overlap: a
candidate may preserve every predicate and numerical value while attaching
them to the wrong temporal context. The subset therefore evaluates
structural semantic alignment beyond token-level coverage.

\noindent\textbf{Informative feedback ablation.} Table~\ref{tab:feedback-ablation}
compares localized corrective feedback with a generic retry instruction
that identifies no semantic discrepancy.

\begin{table}[t]
\centering
\vspace{-1mm}
\caption{\small
True semantic accuracy under informative feedback and an
uninformative retry instruction.
}
\label{tab:feedback-ablation}
\scalebox{0.84}{
\begin{tabular}{lcc}
\toprule
Model & Informative & Uninformative \\
\midrule
Qwen2.5-3B-Instruct & 44.0\% & 17.5\% \\
Qwen2.5-7B-Instruct & 76.0\% & 50.0\% \\
Qwen3.8-27B         & 98.0\% & 88.0\% \\
Qwen3.8-Max         & 99.5\% & 89.9\% \\
\bottomrule
\end{tabular}}
\end{table}

A generic retry yields lower accuracy than localized feedback in all four
conditions. The observed differences are 26.5 and 26.0 percentage points
for the 3B and 7B translators and 10.0 and 9.6 points for the two larger
models. These results indicate that localized semantic content contributes
beyond the opportunity for another decoding attempt. This ablation uses a
dedicated subset, so its values are not interchangeable with the 500-case
main results.

\noindent\textbf{Acceptance noise ablation.} We inject a 10\%
false-positive rate into the acceptance signal. Table~\ref{tab:acceptance-noise}
separates apparent convergence from externally judged accuracy because a
false acceptance ends the loop even when a semantic mismatch remains.

\begin{table}[t]
\centering
\vspace{-1mm}
\caption{\small
Effect of 10\% false-positive acceptance noise. Baseline and noisy
true accuracy are externally evaluated.
}
\label{tab:acceptance-noise}
\scalebox{0.78}{
\begin{tabular}{lccc}
\toprule
Model
& Baseline true Acc.
& Apparent convergence
& Noisy true Acc. \\
\midrule
Qwen2.5-3B-Instruct & 44.0\% & 83.0\%  & 37.0\% \\
Qwen2.5-7B-Instruct & 76.0\% & 92.0\%  & 62.5\% \\
Qwen3.8-27B         & 98.0\% & 99.5\%  & 96.0\% \\
Qwen3.8-Max         & 99.5\% & 100.0\% & 97.4\% \\
\bottomrule
\end{tabular}}
\end{table}

Noise raises apparent convergence while lowering semantic accuracy.
For Qwen2.5-7B, the former reaches 92.0\% as the latter falls from
76.0\% to 62.5\%. The loss is larger for the 3B and 7B translators
(7.0 and 13.5 points) than for the two stronger models (2.0 and 2.1
points), consistent with weaker translators presenting more incorrect
candidates that can be accepted prematurely. The results show that
acceptance quality directly affects whether iteration produces genuine
correction or premature termination.

\section{Human Evaluation and Interaction Cases}
\label{app:human-evaluation}

\noindent\textbf{Expert audit of automatic acceptance.} Domain experts
independently reviewed 634 outputs accepted by the automatic loop. A
case counted as correct only when its complete STL semantics matched
the intended requirement. Experts confirmed 629 cases and rejected five.

Acceptance precision is \(629/634=99.2\%\). The corresponding
false-acceptance rate is \(5/634=0.8\%\), and the
Wilson 95\% confidence interval for acceptance precision is
\([98.2\%,99.7\%]\). Table~\ref{tab:acceptance-audit} reports the
domain-wise results.

\begin{table}[t]
\centering
\vspace{-1mm}
\caption{\small
Expert audit of automatically accepted deliveries.
}
\label{tab:acceptance-audit}
\scalebox{0.88}{
\begin{tabular}{lccc}
\toprule
Domain & Accepted & Confirmed & Precision \\
\midrule
ADAS       & 146 & 145 & 99.3\% \\
Drone      & 146 & 144 & 98.6\% \\
Smart Home & 138 & 137 & 99.3\% \\
AGV        & 137 & 136 & 99.3\% \\
Nested     & 67  & 67  & 100.0\% \\
\midrule
Total      & 634 & 629 & 99.2\% \\
\bottomrule
\end{tabular}}
\end{table}

Because the audit conditions on automatic acceptance, it estimates
acceptance precision and does not measure agreement on an arbitrary
sample. Domain-level precision ranges from 98.6\% to 100.0\%, with the
five false acceptances distributed across four application domains. The
confidence interval should therefore accompany the point estimate when
characterizing reliability beyond the audited outputs.

\noindent\textbf{Human-in-the-loop evaluation.} We replace the
simulated user with human participants in this evaluation. Participants
provide 40 natural-language requirements, each of which is evaluated
with all seven models, yielding 280 model--requirement sessions. In each
session, a participant inspects the semantic observation of the current
STL candidate and either accepts it or provides a natural-language
correction. A correction is counted as feedback only when it informs a
subsequent translation attempt; a rejection after the final translation
round is excluded from this measure.

Of the 280 sessions, 202 terminate with participant acceptance, yielding
an acceptance rate of 72.1\%. Participants provide 1,010 NL inputs in
total, comprising 280 initial requirements and 730 subsequent feedback
messages that inform further translation attempts.
Table~\ref{tab:human-loop} reports the model-wise acceptance rates and
interaction costs.

\begin{table*}[!b]
\centering
\vspace{-1mm}
\caption{\small
Real-user evaluation on 40 shared requirements per model. Feedback
counts participant-written corrections used in a subsequent translation
attempt; final-round rejections are excluded.
}
\label{tab:human-loop}
\scalebox{0.88}{
\begin{tabular}{lcccc}
\toprule
Model
& Accepted
& Acceptance rate
& Feedback
& FB/session
\\
\midrule
Qwen2.5-0.5B-Instruct & 3/40  & 7.5\%   & 297 & 7.43 \\
Qwen2.5-3B-Instruct   & 20/40 & 50.0\%  & 182 & 4.55 \\
Qwen2.5-7B-Instruct   & 25/40 & 62.5\%  & 149 & 3.73 \\
Qwen3.8-27B           & 40/40 & 100.0\% & 12  & 0.30 \\
Qwen3.6-Flash         & 38/40 & 95.0\%  & 28  & 0.70 \\
Qwen3.7-Max           & 37/40 & 92.5\%  & 48  & 1.20 \\
Qwen3.8-Max           & 39/40 & 97.5\%  & 14  & 0.35 \\
\bottomrule
\end{tabular}}
\end{table*}

The 72.1\% participant acceptance rate is distinct from the 99.2\%
automatic acceptance precision. The former measures whether a participant
and translator reach an acceptable specification within the interaction
budget; the latter estimates correctness only among outputs accepted by
the automated judge. Model-wise results also show substantial variation in interaction
cost. Qwen2.5-0.5B-Instruct reaches acceptance in 3 of 40 sessions and
requires 297 feedback messages. The four larger models collectively
reach acceptance in 154 of 160 sessions with 102 feedback messages. The
40 shared requirements support a comparison of these broad trends, although
they do not establish a fine-grained ranking among the larger models.

\noindent\textbf{Representative execution traces.}
To complement the two traces presented in Section~\ref{sec:protocol}
(using Qwen2.5-3B-Instruct), ten additional cases from the
500-instance evaluation are presented below: two cases each for the
0.5B, 3B, 7B, 27B, and Flash translators. Each trace reports the
natural-language requirement, translator candidates $\varphi_t$,
back-translations $\hat{x}_t$, and selected
logged feedback $\delta_t$ from the simulated user. Repeated rounds and
long feedback messages are condensed without changing their reported
outcome. Here, \textsc{accept} records the operational decision made by
the acceptance mechanism; independent correctness is assessed separately
by the expert audit above.

\vspace{2mm}
\noindent\textit{Qwen2.5-0.5B-Instruct and Qwen2.5-3B-Instruct.}
\vspace{1mm}

\noindent\textbf{Connective confusion (0.5B, 7 rounds $\to$ \textsc{accept}).}

\noindent\textit{Requirement:} ``At all times either the service brake
or the spring-applied parking brake must be available.''

\noindent\textit{Round~1.}
$\varphi_0=\mathbf{G}_{[0,\infty)}
(\texttt{service\_brake}\wedge\texttt{parking\_brake})$.
$\hat{x}_0$: ``At all times, both the service brake and the
spring-applied parking brake are available.''
$\delta_0$: ``I meant that at least one must be available: either the
service brake or the spring-applied parking brake. The restatement
incorrectly requires both to be available.''

\noindent\textit{Round~2.} The translator emits an unparseable output;
the scaffold mechanism recovers and re-enters the loop.

\noindent\textit{Rounds~3--6.}
$\varphi_t=\mathbf{G}_{[0,\infty)}
(\texttt{service\_brake}\wedge\texttt{parking\_brake})$ in every
round. $\hat{x}_t$: ``At all times, both the service brake and the
spring-applied parking brake shall be available.''
$\delta_4$: ``I meant `either the service brake or the spring-applied
parking brake,' not `both.' At all times, at least one of the two
brakes must be available; they do not both need to be available
simultaneously.''

\noindent\textit{Round~7.}
$\varphi_6=\mathbf{G}_{[0,\infty)}
(\texttt{service\_brake}\vee\texttt{parking\_brake})$.
$\hat{x}_6$: ``At all times, either the service brake is available or
the spring-applied parking brake is available.''
The acceptance mechanism returns \textsc{accept}. A single-token
$\wedge\!\to\!\vee$ substitution requires six \textsc{revise}
decisions and one scaffold recovery.

\noindent\textbf{Unreachable time bound (0.5B, 10 rounds $\to$ unresolved).}

\noindent\textit{Requirement:} ``Keep the laundry-room exhaust fan
running throughout the next 240 seconds.''

\noindent\textit{Round~1.}
$\varphi_0=\mathbf{G}_{[0,\infty)}
(\texttt{fan\_running}\wedge\texttt{fan\_running})$.
$\hat{x}_0$: ``At all times, `laundry room exhaust fan running' is
active and `laundry room exhaust fan running' is active.''
$\delta_0$: ``I mean throughout the next two hundred forty seconds, not
at all times. The laundry room exhaust fan must keep running throughout
that window, but I do not require it to keep running beyond it.''

\noindent\textit{Rounds~2--10.}
Every candidate is identical to $\varphi_0$ (round~5 triggers a parse
failure that the scaffold recovers).
$\delta_4$: ``I meant `throughout the next 240 seconds,' not `At all
times.' I require the laundry-room exhaust fan to keep running only
within that window, not indefinitely.''
The protocol returns $\mathrm{unresolved}$ after $K=10$. Two errors
survive every round: the bound remains $[0,\infty)$, although the
requirement specifies $[0,240]$, and the predicate remains conjoined with
itself. In this
case, the 0.5B translator does not modify the numerical interval despite
repeated feedback.

\noindent\textbf{Temporal anchoring repair (3B, 3 rounds $\to$ \textsc{accept}).}

\noindent\textit{Requirement:} ``Any time the vehicle reaches the
parking approach zone, the parking planner must become active within 2
seconds.''

\noindent\textit{Round~1.}
$\varphi_0=\mathbf{G}_{[0,2]}
(\texttt{at\_zone}\!\to\!\texttt{planner\_active})$.
$\hat{x}_0$: ``Throughout the first 2 seconds ([0,\,2], both endpoints
included), whenever vehicle at parking approach zone is active, parking
planner active must hold.''
$\delta_0$: ``I did not mean `Throughout the first 2 seconds' from time
0. I meant that each time the vehicle reaches the parking approach
zone, the parking planner must become active within 2 seconds of that
event, not immediately or continuously.''

\noindent\textit{Round~2.}
$\varphi_1=\mathbf{F}_{[0,2]}
(\texttt{at\_zone}\!\to\!\texttt{planner\_active})$.
$\hat{x}_1$: ``At some point within the closed interval [0,\,2] seconds
from time 0, without being triggered by any event, if vehicle at
parking approach zone is active, then parking planner active holds.''
$\delta_1$: ``I meant the obligation to be triggered any time the
vehicle reaches the parking approach zone, not once from time 0. For
every such arrival, the parking planner must become active within 2
seconds of that arrival.''

\noindent\textit{Round~3.}
$\varphi_2=\mathbf{G}_{[0,\infty)}
(\texttt{at\_zone}\!\to\!
\mathbf{F}_{[0,2]}\,\texttt{planner\_active})$.
$\hat{x}_2$: ``Whenever vehicle at parking approach zone is active,
parking planner active must become true within [0,\,2] seconds,
including both endpoints.''
The acceptance mechanism returns \textsc{accept}. The repair moves the
2-second deadline from the outer temporal operator to the inner
eventuality.

\noindent\textbf{Temporal anchoring plateau (3B, 10 rounds $\to$ unresolved).}

\noindent\textit{Requirement:} ``Any time the route becomes
unavailable, the route-guidance fallback must become active within 3
seconds.''

\noindent\textit{Round~1.}
$\varphi_0=\mathbf{G}_{[0,3]}
(\texttt{route\_unavail}\!\to\!
\texttt{fallback\_active})$.
$\hat{x}_0$: ``Throughout the first 3 seconds ([0,\,3], both endpoints
included), whenever route unavailable is active, route guidance
fallback must be active.''
$\delta_0$: ``I meant `within 3 seconds' to be measured from each time
the route becomes unavailable, not from a fixed starting time. I
require the route-guidance fallback to become active by that deadline,
not necessarily immediately.''

\noindent\textit{Rounds~2--10.}
Every candidate is identical to $\varphi_0$; the 3-second interval
never leaves the outer operator.
$\delta_7$: ``I meant `within 3 seconds' to be measured from each time
the route becomes unavailable, not from the start. I meant this to
apply `Any time,' not only `Throughout the first 3 seconds.'\,''
The protocol returns $\mathrm{unresolved}$ after $K=10$. The intended
structure matches the previous case,
$\mathbf{G}_{[0,\infty)}(\cdot\!\to\!\mathbf{F}_{[0,3]}\cdot)$,
yet the 3B translator settles on a fixed point. The same error type is
repaired for one input and persistent for another.

\vspace{2mm}
\noindent\textit{Qwen2.5-7B-Instruct and Qwen3.8-27B.}
\vspace{1mm}

\noindent\textbf{Incremental structural assembly (7B, 7 rounds $\to$ \textsc{accept}).}

\noindent\textit{Requirement:} ``Whenever the driver presses the hazard
switch or an automatic emergency call is placed, the hazard lights must
start flashing within 0.5 seconds.''

\noindent\textit{Round~1.}
$\varphi_0=\mathbf{G}_{[0,\infty)}
(\texttt{switch}\vee\texttt{call})$.
$\hat{x}_0$: ``At all times, `driver pressed hazard switch' is active
or `automatic emergency call placed' is active.''
$\delta_0$: ``I did not mean that either event must be active at all
times. I meant either event to trigger the missing obligation: the
hazard lights must start flashing within 0.5 seconds, with that deadline
measured from the triggering event.''

\noindent\textit{Round~2.}
$\varphi_1=\mathbf{G}_{[0,\infty)}
(\mathbf{F}_{[0,0.5]}\texttt{flash}\vee
 \mathbf{F}_{[0,0.5]}\texttt{flash})$.
The candidate adds the response but drops both triggers. $\delta_1$
restates that each response must be conditioned on the corresponding
event.

\noindent\textit{Round~3.}
$\varphi_2=\mathbf{G}_{[0,\infty)}
((\texttt{switch}\vee\texttt{call})\wedge
 \mathbf{F}_{[0,0.5]}\texttt{flash})$.
$\hat{x}_2$ requires both the trigger disjunction and the response at
all times; the acceptance mechanism returns \textsc{revise}, and the
simulated user provides another correction.

\noindent\textit{Round~4.}
$\varphi_3=\mathbf{G}_{[0,\infty)}
((\texttt{switch}\!\to\!\mathbf{F}_{[0,0.5]}\texttt{flash})
 \vee(\texttt{call}\!\to\!\mathbf{F}_{[0,0.5]}\texttt{flash}))$.
$\hat{x}_3$: ``At all times, at least one of the two trigger--response
conditions must hold.'' The pairs are correct, but the disjunction can
satisfy the formula while one obligation is violated.

\noindent\textit{Rounds~5--6.} The translator retains the
$\vee$-joined pair. The simulated user continues to require both
conditional obligations.

\noindent\textit{Round~7.}
$\varphi_6=(\texttt{switch}\!\to\!
\mathbf{F}_{[0,0.5]}\texttt{flash})\wedge
(\texttt{call}\!\to\!
\mathbf{F}_{[0,0.5]}\texttt{flash})$.
$\hat{x}_6$ states that both trigger--response rules must hold. The
acceptance mechanism returns \textsc{accept}. The interaction
progressively assembles
the trigger, response, implication, and conjunction structure.

\noindent\textbf{Structural stagnation (7B, 10 rounds $\to$ unresolved).}

\noindent\textit{Requirement:} ``For the next 60 seconds, a rear-path
result must be updated within 10 seconds whenever the reverse-motion
inhibit is active.''

\noindent\textit{Round~1.}
$\varphi_0=\mathbf{F}_{[0,60]}
(\texttt{inhibit}\wedge\mathbf{F}_{[0,10]}\texttt{updated})$.
$\hat{x}_0$: ``At some point within [0,\,60] seconds, reverse-motion
inhibit is active and a rear-path result is updated within 10 seconds.''
$\delta_0$: ``I meant the obligation to apply whenever the
reverse-motion inhibit is active throughout the 60 seconds, not just at
one point. Each 10-second deadline must run from the instant the inhibit
is active. I did not require the inhibit to become active at all.''

\noindent\textit{Round~2.}
$\varphi_1=\mathbf{G}_{[0,60]}(\texttt{inhibit}\wedge
(\mathbf{F}_{[0,10]}\texttt{updated}\;
 \mathbf{U}_{[0,\infty)}\;\texttt{inhibit}))$.
$\hat{x}_1$ requires the inhibit to hold throughout the 60-second
window and nests the update inside an Until clause.
$\delta_1$: ``I did not require the reverse-motion inhibit to remain
active throughout that period. I require a rear-path result update
within each triggered deadline.''

\noindent\textit{Rounds~3--10.} The translator reproduces
$\varphi_1$ unchanged after every targeted correction. The protocol
returns $\mathrm{unresolved}$ after $K=10$. The predicates and intervals
are retained, while the required implication remains absent.

\noindent\textbf{Reach-and-maintain via Until (27B, 8 rounds $\to$ \textsc{accept}).}

\noindent\textit{Requirement:} ``The drone must reach the cruise
altitude band within 40 seconds and keep the altitude between 78 and
82\,m for the following 5 minutes.''

\noindent\textit{Round~1.}
$\varphi_0=\mathbf{F}_{[0,40]}
((78\!\leq\!\texttt{alt}\!\leq\!82)\wedge
 \mathbf{G}_{[0,300]}(78\!\leq\!\texttt{alt}\!\leq\!82))$.
$\hat{x}_0$ places reaching the band and the 300-second invariance
inside one future state.
$\delta_0$: ``The drone may be outside the altitude band before it
reaches between 78 and 82 meters within 40 seconds. The 5 minutes must
begin when the drone reaches that band, not at time 0.''

\noindent\textit{Rounds~2--7.} The candidates alternate among a global
implication, a conjunction of reach and maintenance, and the original
$\mathbf{F}_{[0,40]}$ structure. The user repeatedly specifies that
reaching is an achievement obligation and that the maintenance interval
starts at the achieved state.

\noindent\textit{Round~8.}
$\varphi_7=\neg(78\!\leq\!\texttt{alt}\!\leq\!82)\;
\mathbf{U}_{[0,40]}\;
((78\!\leq\!\texttt{alt}\!\leq\!82)\wedge
 \mathbf{G}_{[0,300]}(78\!\leq\!\texttt{alt}\!\leq\!82))$.
$\hat{x}_7$ states that the altitude remains outside the band until a
point within 40 seconds where the band holds and is then maintained for
300 seconds. The acceptance mechanism returns \textsc{accept}. The Until
operator sequences the reaching and maintenance phases.

\noindent\textbf{Periodic recurrence gap (27B, 10 rounds $\to$ unresolved).}

\noindent\textit{Requirement:} ``Starting from midnight, the cleaning
robot must perform housework and cleaning every day from 10:00 AM to
10:30 AM. After finishing, it must wash and store its tools, then charge
itself while keeping Bluetooth enabled and WiFi connected.''

\noindent\textit{Round~1.}
$\varphi_0$ contains
$\mathbf{G}_{[0,\infty)}(\texttt{housework}\wedge\texttt{cleaning})$
and two implications for tool handling and connectivity.
$\hat{x}_0$: ``At all times, the cleaning robot is actively performing
housework and cleaning; whenever cleaning finishes, it washes and stores
its tools; whenever it charges, Bluetooth and WiFi remain active.''
$\delta_0$: ``I meant cleaning to occur every day from 10:00 AM to
10:30 AM, not at all times. After washing and storing its tools, the
robot must then charge itself. Charging must be an ordered obligation.''

\noindent\textit{Round~2.}
The schedule becomes
$\mathbf{G}_{[36000,36180]}
(\texttt{housework}\wedge\texttt{cleaning})$; the remaining formula
adds an Until relation from tool handling to charging.
$\hat{x}_1$ interprets $[36000,36180]$ as a single 10:00--10:03 AM
window. The simulated user reiterates ``every day from 10:00 AM to
10:30 AM.''

\noindent\textit{Rounds~3--7.} The same single-window encoding and its
incorrect end time persist.

\noindent\textit{Rounds~8--10.} The schedule reverts to
$\mathbf{G}_{[0,\infty)}(\texttt{housework}\wedge\texttt{cleaning})$,
while the sequential post-cleaning clauses improve. The daily
recurrence is never represented. The protocol returns
$\mathrm{unresolved}$ after $K=10$. This trace illustrates a periodic
schedule that the bounded STL fragment and translation pipeline used in
this evaluation do not represent directly.

\vspace{2mm}
\noindent\textit{Qwen3.6-Flash.}
\vspace{1mm}

\noindent\textbf{Multi-conjunct progressive repair (Flash, 6 rounds $\to$ \textsc{accept}).}

\noindent\textit{Requirement:} ``Starting from midnight, the irrigation
system must water the garden for exactly 15 minutes beginning at 6:00
AM. After watering, soil moisture must exceed 60\% within 5 minutes. If
it drops below 40\% again before noon, the system must trigger a second
10-minute watering cycle. Water pressure must remain between 200 and
400\,kPa throughout all watering.''

\noindent\textit{Round~1.}
$\varphi_0$ combines
$\mathbf{G}_{[21600,22500]}\texttt{watering}$,
$\mathbf{G}_{[0,\infty)}\mathbf{F}_{[22500,22800]}
(\texttt{moisture}>0.6)$, a globally scoped low-moisture trigger whose
response is $\mathbf{G}_{[0,600]}\texttt{watering}$, and a pressure
constraint on \texttt{water\_pressure}.
$\hat{x}_0$ interprets the moisture deadline as a moving obligation and
the second cycle as watering during the first 600 seconds.
$\delta_0$: ``I meant water pressure to remain between 200 and 400 kPa.
I meant the soil moisture sensor to read above 60\% within 5 minutes
after watering, not to satisfy a moving deadline at every moment. I
meant a drop below 40\% again before 12:00 PM to trigger a second
watering cycle of 10 minutes following that drop.''

\noindent\textit{Round~2.} The global wrapper is removed from the
post-watering moisture deadline. The second-cycle clause becomes an
implication from $\mathbf{F}_{[22500,43200]}(\texttt{moisture}<0.4)$
to $\mathbf{G}_{[0,600]}\texttt{watering}$.
$\delta_1$: ``I need that cycle to follow the low-moisture trigger, not
require watering throughout the first 600 seconds from midnight.''

\noindent\textit{Round~3.} The translator introduces
$\mathbf{G}_{[0,43200]}((\texttt{moisture}<0.4)\!\to\!
\mathbf{F}_{[0,600]}\mathbf{G}_{[0,600]}\texttt{watering})$.
The user rejects the added 600-second start deadline and clarifies that
the trigger applies only after the initial watering.

\noindent\textit{Rounds~4--5.} The added start deadline is removed, and
the trigger scope is narrowed first to $[0,43200]$ and then to
$[22500,43200]$. The remaining discrepancy is the pressure unit.
$\delta_4$: ``I meant water pressure to remain between 200 and 400 kPa,
not between 200 and 400 in Pa. Preserve kPa or correctly convert the
bounds when using Pa.''

\noindent\textit{Round~6.}
$\varphi_5$ retains the four corrected clauses and replaces the pressure
signal with \texttt{water\_pressure\_kpa}. $\hat{x}_5$ preserves the
6:00--6:15 AM watering window, post-watering moisture deadline,
conditional second cycle, and pressure range. The acceptance mechanism
returns \textsc{accept}. In this trace, each retained correction remains
present in the accepted candidate.

\noindent\textbf{False mismatch from back-translation (Flash, 10 rounds $\to$ unresolved).}

\noindent\textit{Requirement:} ``Whenever solar generation is greater
than household demand, raise the battery charging command to at least
30 percent within 10 seconds.''

\noindent\textit{Round~1.}
$\varphi_0=\mathbf{G}_{[0,\infty)}
(\texttt{solar}\!>\!\texttt{demand}\!\to\!
\mathbf{F}_{[0,10]}(\texttt{battery\_cmd}\!\geq\!0.3))$.
The candidate is semantically correct. $\hat{x}_0$: ``Whenever solar
generation is not equal to household demand, the battery charging
command must reach at least 0.3 within [0,\,10] seconds.''
$\delta_0$: ``I meant the trigger to be `solar generation is greater
than household demand,' not `solar generation is not equal to household
demand.' I do not want this obligation to apply when solar generation
is less than household demand.''

\noindent\textit{Rounds~2--10.} The translator reproduces the same
correct $\varphi_0$ in every round. The back-translator renders $>$ as
``not equal to'' every time, so the simulated user repeats the same
correction. The acceptance mechanism records a semantic mismatch and
returns \textsc{revise}. The protocol returns $\mathrm{unresolved}$ after
$K=10$. In this logged case, the failure arises in $B$: the forward
output is correct from round~1, while the distorted observation prevents
acceptance.

\noindent\textbf{Interface-level failure modes.} The audit identifies
two additional interface errors. In one false rejection, a semantically
equivalent unit expression is treated as a surface mismatch. In one
false acceptance, a conjunction--disjunction scope ambiguity changes the
logical dependency but passes the calibration panel. Together with the
comparison-rendering error above, these cases motivate stronger unit
normalization, explicit Boolean-scope auditing, and improved comparison
rendering.

\section{Limitations}
Refinement gains depend on the translator's observed ability to incorporate feedback. In this evaluation, the 0.5B model gains only 1.0 percentage point despite frequent interaction, and full-formula regeneration can modify previously correct subformulas. The current results therefore characterize the seven evaluated translators under a ten-round budget; they do not establish a model-independent capability threshold.

A further limitation concerns predicate grounding. Descriptions such as ``a blue ball'' do not by themselves define an executable predicate. Deployment requires choices about sensing modalities, representation spaces, decision thresholds, object identity, and robustness to environmental variation. A predicate-grounding layer must connect these choices to the formal specification, with additional clarification when the intended grounding is not unique. External validity is also limited by the focus on STL, four cyber-physical-system domains, and 280 human-interaction sessions.

\section{Future Directions and Extensions}
\label{app:future-work}

We hypothesize that semantic observation and discrepancy-driven refinement could extend beyond STL to other executable formal languages, including linear temporal logics, planning specifications, program contracts, and workflow descriptions. This hypothesis is not evaluated here. Such extensions would require language-specific observation mechanisms and separately calibrated acceptance criteria.

A second direction is to study semantic acceptance together with formal verification, runtime monitoring, and controller synthesis. Counterexamples, execution traces, or behavioral trajectories could provide additional feedback signals after intent-level review. Evaluating these signals within a unified closed-loop framework remains future work.

\end{document}